# Learning Graphs with Monotone Topology Properties and Multiple Connected Components

Eduardo Pavez, Hilmi E. Egilmez and Antonio Ortega

*Abstract*—Recent papers have formulated the problem of learning graphs from data as an inverse covariance estimation with graph Laplacian constraints. While such problems are convex, existing methods cannot guarantee that solutions will have specific graph topology properties (e.g., being a tree or $k$-partite), which are desirable for some applications. In fact, the problem of learning a graph with given topology properties, e.g., finding the $k$-partite graph that best matches the data, is in general non-convex. In this paper, we develop novel theoretical results that provide performance guarantees for an approach to solve these problems. Our solution decomposes this graph learning problem into two sub-problems, for which efficient solutions are known. Specifically, a graph topology inference (GTI) step is employed to select a feasible graph topology, i.e., one having the desired topology property. Then, a graph weight estimation (GWE) step is performed by solving a generalized graph Laplacian estimation problem, where edges are constrained by the topology found in the GTI step. Our main result is a bound on the error of the GWE step as a function of the error in the GTI step. This error bound indicates that the GTI step should be solved using an algorithm that approximates the similarity matrix (which in general corresponds to a complete weighted graph) by another matrix whose entries have been thresholded to zero to have the desired type of graph coloring. The GTI stage can leverage existing methods (e.g., state of the art approaches for graph coloring) which are typically based on minimizing the total number of removed edges. Since the GWE stage is formulated as an inverse covariance estimation problem with linear constraints, it can be solved using existing convex optimization methods. We demonstrate that our two step approach can achieve good results for both synthetic and texture image data.

## I. INTRODUCTION

**G**RAPHS are mathematical structures, consisting of nodes (vertices) and links (edges), used in various fields to represent and analyze signals. Particularly, in signal processing and machine learning, graphs have been extensively used for modeling high dimensional datasets, where graphs' nodes represent objects of interest and the edges with designated weights encode pairwise relations between them. Applications of such models include transformation [1], filtering [2], [3] and sampling of signals defined on graphs [4], as well as clustering [5], [6], semi-supervised learning [7], dynamical systems [8], [9], and network-oriented data problems [10], [11].

In many cases, datasets consist of an unstructured list of samples, and the underlying graph information (representing the structural relations between samples) is unknown. A fundamental problem is to characterize the hidden structure (i.e., set of edges) and affinity relations (weights) between the entries of a dataset. Effective graph-based modeling depends on the quality and type of graph. Thus, it is crucial to develop graph learning methods to optimize the graph topology and the link weights.

Graphs can be represented by different types of matrices (such as adjacency and graph Laplacian matrices), whose nonzero entries correspond to edges in the graph. The choice of a matrix type may depend on modeling assumptions, properties of the desired graph, applications and theoretical requirements. In this work, we will consider generalized graph Laplacian matrices (GGL) [12], which are positive definite matrices with nonpositive off-diagonal entries used to represent undirected graphs with non-negative weights and self-loops. GGL matrices arise in statistics as precision matrices of attractive Gaussian Markov random fields [13]. They have also been applied to image and video coding [1], [14], in algorithms for graph sparsification [15], [16], for solving linear systems [17], and for analyzing electrical circuits [18]. Several theoretical properties of GGL matrices and their implications for graph cut based algorithms are discussed in [19], [20].

In the context of graph signal processing [21], [22], algorithms have been proposed that are restricted to (or operate more efficiently on) graphs with specific topology properties. For example, bipartite graphs are required for perfect reconstruction two channel filter banks [23], $M$-block cyclic graphs are required for designing $M$-channel filter banks [24], [25]. Tree structured graphs have been used to design multiresolution transforms [26], [27] and sampling algorithms [28]. Decomposition of a graph into connected sub-graphs can also help constructing new multiresolution analysis for graph signals [29], analyze performance of spectral clustering [30] and can be useful to scale inverse covariance estimation algorithms [31]. A bipartite graph with multiple connected components is used in [32] for co-clustering. Sparsity, which is often a desirable graph property, can be achieved by tuning $\ell_1$-regularization parameters in the graph learning optimization [33]. In contrast, none of the aforementioned topology properties (e.g., bipartition or a certain number of connected components) can be directly achieved by adding penalty functions to an existing graph learning problem.

In this paper, we consider a graph learning problem where the goal is to find a graph that best fits the data, while having a specific topology property. For instance, if we are interested in the family of tree structured graphs, we will need to find the best tree topology and weights for the data. The problem of

Authors are with the Department of Electrical Engineering, University of Southern California, Los Angeles, CA, 90089 USA. Corresponding author e-mail: pavezcar@usc.edu. This paper was funded in part by NSF under grant: CCF-1410009.

graph learning with topology properties naturally generalizes the work from [33], [34], [13], where the graph structure was either fixed or unknown.

Specifically, given data and a desired topology property our goal is to solve the following problem.

**Problem 1.** Let $\mathbb{L}_n^+$ be the set of $n \times n$ positive definite GGL matrices, $\mathbf{S}$ be a similarity (e.g., covariance or kernel) matrix computed from data, and let $\mathcal{F}$ be a family of graphs with a certain property. The goal is to find a GGL matrix $\mathbf{L}$ of a weighted graph $\mathcal{G}(\mathbf{L})$, that solves:

$$\min_{\mathbf{L}} - \log \det(\mathbf{L}) + \operatorname{tr}(\mathbf{SL}), \text{ s.t. } \mathbf{L} \in \mathbb{L}_n^+, \mathcal{G}(\mathbf{L}) \in \mathcal{F}.$$

The objective function in Problem 1 is convex [35], and its optimization can be viewed from a probabilistic perspective as inverse covariance estimation [36], [37], [33] or as matrix approximation using a log-determinant Bregman divergence [38]. However, the constraint set $\{\mathbf{L} \in \mathbb{L}_n^+ : \mathcal{G}(\mathbf{L}) \in \mathcal{F}\}$ is not convex for many graph families (see Proposition 1), making Problem 1 non-convex, and thus intractable.

We propose a two step procedure (Algorithm 1) to find an approximate solution for Problem 1. The first step is graph topology inference (GTI), where we can use any algorithm that returns a graph with the desired topology property, in particular algorithms that operate *directly* on the similarity matrix $\mathbf{S}$, i.e., without attempting to find its inverse. The second step is graph weight estimation (GWE), where we solve the convex problem of estimating a GGL matrix with an edge set contained within the topology found in the first step, using for example the algorithm from our previous work [33].

Our main contributions are a set of theoretical results that allow us to derive performance guarantees for the aforementioned two-step graph learning framework.

First, in Section IV we analyze a weighted $\ell_1$-regularized GGL estimation problem:

**Problem 2.** Given an arbitrary regularization matrix $\boldsymbol{\Gamma} = (\gamma_{ij})$, and a similarity matrix $\mathbf{S}$, find $\mathbf{L}$ that solves

$$\min_{\mathbf{L}} - \log \det(\mathbf{L}) + \operatorname{tr}(\mathbf{SL}) + \sum_{i \neq j} \gamma_{ij} |l_{ij}|, \text{ s.t. } \mathbf{L} \in \mathbb{L}_n^+.$$

This problem serves as a convex relaxation of Problem 1. In particular, we characterize topology properties of the solution of Problem 2 as a function of $\boldsymbol{\Gamma}$ and $\mathbf{S}$. This set of results are crucial in the derivation of the proposed algorithm and its performance guarantees.

Second, in Section V we show that the error between the solution of Problem 1 and the output of the GWE step in Algorithm 1 can be bounded by an error measure that depends on the similarity matrix and the topology obatained in the GTI step (Theorem 4 and Lemma 1).

Third, since this error bound is agnostic to how the GTI step was solved, we can leverage existing graph approximation methods, e.g., state of the art graph coloring and tree approximation algorithms, to solve the GTI step. This is because, we can view the GTI step as the problem of taking the complete graph representing the similarity matrix and removing some of its edges until it has the desired topology property. A GTI algorithm that achieves this approximation of the complete similarity graph with lower error, is then guaranteed to provide an error bound for our solution to Problem 1. We illustrate this in Section VI for trees, bipartite graphs, and $k$-sparse connected graphs. For acyclic graphs, we demonstrate that a tighter bound can be achieved (Proposition 4) by leveraging a result in [39] about GGL matrices of trees. With this we show the equivalence between the Chow-Liu algorithm [40] and our proposed method applied to acyclic graphs.

Finally, we derive necessary conditions on the graph families for which the theoretical guarantees of Algorithm 1 hold. One of these conditions is for the graph family of interest to be *monotone*, i.e., closed under edge deletion operations. This is a mild requirement and families satisfying it include acyclic graphs, $k$-partite graphs and $k$-sparse graphs (graphs with at most $k$ edges). We can also handle the non-monotone family of graphs with $J$ connected components, and intersections of a monotone family with the family of graphs with $J$ connected components. This intersection includes other types of non-monotone graph families, such as trees (acyclic connected graphs) and $k$-sparse connected graphs (see Definitions 4 and 5).

The rest of this paper is organized as follows. In Section II, we review the literature of structured and unstructured graph learning methods. In Section III, we introduce the required graph theoretic concepts and present our algorithm. In Section IV, we show our theoretical results on sparsity patterns of the solution of Problem 2. In Section V, we establish the relation between Problems 1 and 2 as well as the error bounds for Algorithm 1. In Section VI, we derive instances of our methods for specific graph families. Finally, numerical validation and conclusions are provided in Sections VII and VIII, respectively.

## II. RELATED WORK

Solving Problem 1 can be viewed as finding an approximate inverse of $\mathbf{S}$ with some type of sparsity pattern. When the similarity matrix is positive definite, Problem 1 is equivalent to minimizing the log-determinant Bregman divergence [38]. Moreover, when the similarity matrix is a sample covariance matrix, Problem 1 can be viewed as maximum likelihood estimator of the inverse covariance matrix of an attractive Gaussian Markov random field (GMRF)[33]. In this section, we review the literature on inverse covariance estimation. In Section II-A, we introduce (weighted) graphical Lasso estimators and discuss their relation with Problem 2. Next, we discuss some studies about sparsity properties of the graphical Lasso and covariance thresholding in Section II-B, which are related to our results of Section IV. We finish with Section II-C discussing papers that study inverse covariance estimators with topology properties.

### A. Estimation of Gaussian Markov random fields

Gaussian Markov random fields (GMRFs) or Gaussian graphical models are characterized by a graph based on the non-zero pattern of the inverse covariance matrix. The edge weights are the entries of the inverse covariance (precision) matrix. A weight with zero value indicates that the corresponding two variables are conditionally independent given the rest. Estimating inverse covariance matrices with a given non-zero pattern constraint is called covariance selection [36]. Graph topology and weights can be found using inverse covariance estimation, for example with the algorithms described in [44], [45]. One of the most widely used methods is the $\ell_1$ regularized Gaussian maximum likelihood estimator, also called graphical Lasso [44], [37], for which several efficient algorithms have been proposed [46], [44], [47].

| Family/Property | Paper | Learning goal | Model | Algorithm | Optimality/Analysis |
|---|---|---|---|---|---|
| Tree | [41] | topology | GMRF | Chow-Liu [40] | consistency |
| | [42] | topology and weights | GMRF | structural graphical Lasso | none |
| | [43] | topology | Ising | graphical Lasso | consistency |
| $k$-connected components | [31] | topology | GMRF | weighted $\ell_1$ graphical Lasso | error bound |

TABLE I: Algorithms for graph learning with topology properties.

Graphical Lasso estimators solve an optimization of the form

$$\min_{\boldsymbol{\Theta} \succ 0} - \log \det(\boldsymbol{\Theta}) + \operatorname{tr}(\mathbf{S}\boldsymbol{\Theta}) + \sum_{i,j} \delta_{ij} |\boldsymbol{\Theta}_{ij}|,$$

with non-negative regularization weights $\delta_{ij}$. The graphical Lasso is an optimization problem over the set of positive definite matrices, while Problems 1 and 2 are defined over the set of positive definite GGL matrices

Attractive GMRFs are a subclass of GMRFs for which the partial correlations are only allowed to be non-negative, so that the corresponding precision matrix is a GGL matrix, having non positive off diagonal entries. Lake and Tenenbaum [48] proposed an inverse covariance estimation problem with a particular type of GGL constraint with all equal and constant self-loop weights. More recently, Slawski and Hein [13] studied the case of arbitrary GGL matrices (Problem 2 without $\ell_1$ regularization term), and show some statistical properties of this estimator, along with an efficient algorithm.

In our recent work [33], we generalized the algorithm from [13] to allow for structural (known topology) constraints and $\ell_1$-regularization. We considered different types of attractive GMRFs obtained by using several types of Laplacian matrices, namely generalized graph Laplacian (GGL), diagonally dominant GGL and combinatorial graph Laplacian matrices. These algorithms are useful when the graph topology is available, e.g., images or video. However, there are cases where the graph does not have a known topology and instead is required to have a certain property such as being connected or bipartite (refer to the Introduction for examples). In these cases, new algorithms need to be developed to chose a graph structure within that family. The algorithm proposed in this paper for graph learning with topology properties complements our previous work by providing efficient ways of choosing the graph topology constraints. Moreover, we can use several solvers for solving the GWE step of our algorithm [33], [34], [49], [13].

### B. Characterizations of graphical model structures

The results from Sections IV and V are based on the analysis of the properties of the solution of Problem 2. Similar characterizations are available for the graphical Lasso [44], [47] and its weighted variant [31]. For example, it has been shown that the graphical Lasso solution and a graph obtained by thresholding the entries of the covariance matrix have the same block diagonal structure [50], [51], i.e., the same connected components. In Theorem 2, we show an equivalent result for Problem 2 with a different covariance thresholding graph.

In [31] the $\ell_1$ regularization parameters are designed to control the covariance thresholding graph, such that the solution of the weighted graphical Lasso has a certain number of connected components. Our work follows a similar strategy to learn graphs with topology properties, but we focus on learning graphs represented by GGL matrices. Our approach not only guarantees a decomposition into connected components, but also allows analyzing other graph properties.

For certain types of sparse graphs obtained by solving graphical Lasso with high regularization, it was shown in [52] that covariance thresholding and graphical Lasso have the same graph structure. However, the equivalence conditions are hard to check, except for acyclic graphs as shown in [53]. In [54], a sequel to [52], [53], it is shown that under certain conditions, the graphical Lasso solution for acyclic graphs has a closed form expression. The same formula from [54] was previously derived in [39] but in the context of tree structured M-matrices[1]. In this paper, we show that covariance thresholding and Problem 2 are equivalent (in graph structure inference) for acyclic graphs. We also show that by applying the formula from [39] we can obtain a much shorter proof than the one in [54], but only valid for GGL matrices. In [49] acyclic graphs are studied for the un-regularized GGL estimator. Our results complement those of [49] and are derived for a more general optimization problem.

In Theorem 1, we obtain a characterization of the graph topology of the solution of Problem 2. This is one of the key results of this paper and depends on special properties of GGL matrices. In particular, it allows us to derive proofs for acyclic graphs and graphs with a given number of connected components, similar to those included in the aforementioned papers. However, to the best of our knowledge there are no equivalent results to Theorem 1 for the graphical Lasso or other inverse covariance matrix estimation algorithms.

---

[1]The set of symmetric M-matrices is equal to the set of positive semi-definite GGL matrices.

### C. Estimation of structured GMRFs

Estimation of structured GMRFs can be described as follows: given a GMRF, which is known to have a certain graph topology (e.g., being bipartite), identify the graph structure and/or graph weights from i.i.d. realizations. Identifying the structure of a GMRF is NP-hard for general graph families [55], [56]. The Chow-Liu algorithm [40] was proposed for approximating discrete probability distributions using tree structured random variables. The work of [41] proved that the Chow-Liu algorithm applied to GMRFs is consistent with an exponential rate. We show that our method, when applied to tree structured attractive GMRFs, is equivalent to the Chow-Liu algorithm (see Section VI-A). Other methods that learn tree structured graphs include the structural graphical Lasso [42] and the work of [43], both of which use the graphical Lasso to recover tree structured Ising models.

A line of work in characterization and inference of graphs with certain properties is developed in [57], [55]. They consider random graph models, which include Erdos-Renyi, power-law and small-world random graphs. These algorithms are based on thresholding conditional mutual information and conditional covariances and were shown to be consistent for graph structure estimation.

We summarize the papers for learning graphs with deterministic topology properties in Table I. With respect to these works, our paper has three main differences. First, we focus on finding a graph with a desired property for a given dataset, while most of these works study algorithms that are statistically consistent, i.e., they converge to the true model as the number of observations increase. Second, our error bounds are deterministic, non-asymptotic and are derived only using optimization arguments. Third, our framework includes the graph properties listed in Table I as well as others to be introduced in the next section.

## III. PRELIMINARIES AND PROBLEM STATEMENT

### A. Graph theoretic background

We denote scalars by letters in regular font, while vectors and matrices are written in lower and uppercase bold respectively. Positive definite (PD) and positive semi-definite (PSD) matrices are denoted by $\mathbf{X} \succ 0$ and $\mathbf{X} \succeq 0$, while the inequality notation $\mathbf{X} \geq \mathbf{Y}$ is entry-wise scalar inequality.

A weighted graph is denoted by $\mathcal{G} = (\mathcal{V}, \mathcal{E}, \mathbf{W}, \mathbf{V})$, where $\mathcal{V}$ is a set of vertices (or nodes) of size $|\mathcal{V}| = n$, and $\mathcal{E} \subset \mathcal{V} \times \mathcal{V}$ are the edges (or links). The edge weights are stored in the non-negative matrix $\mathbf{W} = (w_{ij}) \geq 0$, and the vertex weights are stored in $\mathbf{V} = \text{diag}(v_1, v_2, \cdots, v_n)$. We say that there is an edge between vertices $i$ and $j$ if the weight $w_{ij}$ is positive. If there is no edge, then $w_{ij} = 0$. If $v_i \neq 0$ we say that there is a self-loop at node $i$. All graphs are undirected, i.e., $\mathbf{W} = \mathbf{W}^T$. The degree matrix is defined as $\mathbf{D} = \text{diag}(\mathbf{W}\mathbf{1})$ and its diagonal entry $d_i$ is the degree of the $i$-th vertex.

**Definition 1.** [12] The generalized graph Laplacian (GGL) matrix of a graph $\mathcal{G} = (\mathcal{V}, \mathcal{E}, \mathbf{W}, \mathbf{V})$ is defined as

$$\mathbf{L} = \mathbf{V} + \mathbf{D} - \mathbf{W}.$$

With some abuse of notation we will also use $\mathcal{G}(\mathbf{L}) = (\mathcal{V}, \mathcal{E}, \mathbf{L})$, given that $\mathbf{W}$ can be recovered from the off diagonal elements of $\mathbf{L}$ and $\mathbf{V} = \text{diag}(\mathbf{v})$, where $\mathbf{v} = \mathbf{L}\mathbf{1}$.

Positive semi-definite generalized Laplacian matrices are exactly the set of symmetric M-matrices [58]. If $\mathbf{v} \geq \mathbf{0}$ we say the graph is diagonally dominant. If all $v_i = 0$, then we say the graph is simple (no self loops), where $\mathbf{L} = \mathbf{D} - \mathbf{W}$ is called combinatorial Laplacian. We will denote the set of positive definite generalized Laplacian matrices by $\mathbb{L}_n^+$, and set of GGL matrices that do not allow links outside of $\mathcal{E}$ by

$$\mathbb{L}_n^+(\mathcal{E}) = \{\mathbf{L} \in \mathbb{L}_n^+ : l_{ij} = 0, \text{ if } (i,j) \notin \mathcal{E}\}.$$

We will also denote by $\mathbf{L}(\mathcal{E})$ a matrix that keeps only the entries corresponding to an edge set $\mathcal{E}$ and zeroes out the rest, therefore $\mathbf{L} \in \mathbb{L}_n^+(\mathcal{E})$ if and only if $\mathbf{L}(\mathcal{E}) = \mathbf{L}$.

Now we introduce some graph theoretic concepts and graph families used throughout the paper.

**Definition 2.** For a graph with vertex and edge sets $\mathcal{V}$ and $\mathcal{E}$ respectively we define:

- The *neighborhood* of a node $i$ is the subset of vertices $\mathcal{N}_i(\mathcal{E}) = \{j : (i,j) \in \mathcal{E}\}$.
- A *path* from node $i_1 \in \mathcal{V}$ to node $i_t \in \mathcal{V}$ is a sequence of pairs satisfying $\{(i_p, i_{p+1})\}_{p=1}^{t-1} \subset \mathcal{E}$.
- A *cycle* is a path from a node to itself.

**Definition 3.** A graph family (or class) is a set of graphs that have a common property. We denote a generic graph family by $\mathcal{F}$. Some of them include:

- *$k$-sparse* graphs have at most $k$ edges (i.e., $|\mathcal{E}| \leq k$).
- *Bounded degree* graphs have at most $d$ neighbors per node (i.e., $|\mathcal{N}_i| \leq d$ for all $i \in \mathcal{V}$).
- In *connected* graphs, for all $i, j \in \mathcal{V}$, there is a path connecting node $i$ to node $j$.
- Graphs with $k$-*connected components* contain partitions of the vertices $\mathcal{V} = \cup_{i=t}^k \mathcal{S}_t$, such that each sub-graph $\mathcal{G}_t = (\mathcal{V}_t, \mathcal{E}_t)$ is connected. The edges $\mathcal{E}_t \subset \mathcal{E}$ have endpoints in $\mathcal{V}_t$, and no edges connect different connected components.
- *$k$-partite* graphs have partitions of the vertices $\mathcal{V} = \cup_{i=t}^k \mathcal{S}_t$ such that for each $t$ there are no edges within $\mathcal{S}_t$, i.e. for each $(i,j) \in \mathcal{S}_t \times \mathcal{S}_t$ then $(i,j) \notin \mathcal{E}$. The case $k = 2$ corresponds to the family of bipartite graphs.
- *Acyclic* graphs have no cycles. If the graph is also connected it is called a *tree*, otherwise it is called a *forest*.

Graph families can also be characterized by their properties, we are particularly interested in monotone graph families.

**Definition 4.** A graph family (or property) is called *monotone* if it is closed under edge deletion operations. Monotone families include $k$-partite, acyclic, $k$-sparse, and bounded degree graphs.

The following definition contains all graph families required for the theoretical guarantees of Section V.

**Definition 5.** A graph family $\mathcal{F}$ is admissible if one of the following properties hold

- $\mathcal{F}$ is monotone,

- $\mathcal{F}$ is the family of graphs with $k$-connected components,
- $\mathcal{F} = \mathcal{F}_1 \cap \mathcal{F}_2$, where $\mathcal{F}_1$ is a monotone family, and $\mathcal{F}_2$ are the graphs with $k$-connected components. Note that in this case $\mathcal{F}$ is no longer monotone.

### B. Similarity matrix

Let $\mathbf{X} \in \mathbb{R}^{n \times N}$ be a data matrix, denote its $i$-th column by $\mathbf{x}_i$, of dimension $n \times 1$, and its $i$-th row by $\mathbf{x}^i$, of dimension $1 \times N$. Each $\mathbf{x}^i$ is a data point attached to one of the nodes in $\mathcal{V} = \{1, \cdots, n\}$. We use a similarity function $\varphi : \mathbb{R}^N \times \mathbb{R}^N \longrightarrow \mathbb{R}$, that is symmetric $\varphi(\mathbf{a}, \mathbf{b}) = \varphi(\mathbf{b}, \mathbf{a})$ for all $\mathbf{a}, \mathbf{b} \in \mathbb{R}^N$. We define the similarity matrix $\mathbf{S} = (s_{ij})$ between data points at different nodes as $s_{ij} = \varphi(\mathbf{x}^i, \mathbf{x}^j)$. Typical examples of similarity functions are $\varphi(\mathbf{a}, \mathbf{b}) = \frac{1}{N}\langle \mathbf{a}, \mathbf{b} \rangle$ and $\varphi(\mathbf{a}, \mathbf{b}) = \langle \frac{\mathbf{a}}{\|\mathbf{a}\|}, \frac{\mathbf{b}}{\|\mathbf{b}\|} \rangle$ that produce empirical correlation and correlation coefficient matrices, and a Gaussian kernel function $\varphi(\mathbf{a}, \mathbf{b}) = \exp(-\|\mathbf{a} - \mathbf{b}\|^2/\sigma^2)$.

### C. Analysis of Problem 1

We will denote the cost function from Problem 1 by

$$\mathcal{J}(\mathbf{L}) = -\log\det(\mathbf{L}) + \text{tr}(\mathbf{SL}). \quad (1)$$

Even though $\mathcal{J}$ is convex, the constraint set $\{\mathbf{L} \in \mathbb{L}_n^+ : \mathcal{G}(\mathbf{L}) \in \mathcal{F}\}$ is not convex in general. This can be characterized by the following proposition.

**Proposition 1.** *Let $\mathcal{F}$ be any graph family. The constraint set*

$$\{\mathbf{L} \in \mathbb{L}_n^+ : \mathcal{G}(\mathbf{L}) \in \mathcal{F}\}$$

*is convex if and only if $\mathcal{F}$ is closed under edge addition operations.*

*Proof.* Let $\mathbf{L}_1, \mathbf{L}_2$ be arbitrary positive definite GGL matrices, with $\mathcal{G}(\mathbf{L}_1) = (\mathcal{V}, \mathcal{E}_1, \mathbf{L}_1) \in \mathcal{F}$, $\mathcal{G}(\mathbf{L}_2) = (\mathcal{V}, \mathcal{E}_2, \mathbf{L}_2) \in \mathcal{F}$, and pick $\alpha \in (0, 1)$. We have that $\mathbf{L} = \alpha\mathbf{L}_1 + (1-\alpha)\mathbf{L}_2$ is obviously a positive definite GGL matrix. Also, notice that since GGL matrices have non positive off diagonal entries, the convex combination of GGL matrices has a graph given by $\mathcal{G}(\mathbf{L}) = (\mathcal{V}, \mathcal{E}, \mathbf{L})$, with $\mathcal{E} = \mathcal{E}_1 \cup \mathcal{E}_2$, which means that there is an edge in the graph $\mathcal{G}(\mathbf{L})$ if and only if there is either an edge in $\mathcal{G}(\mathbf{L}_1)$ or in $\mathcal{G}(\mathbf{L}_2)$. This implies that for the problem to be convex $\mathcal{F}$ has to be closed under union of edge sets, and therefore closed under edge addition operations. $\quad\square$

When $\mathcal{F}$ is the family of connected graphs, the constraint is convex, i.e., adding an edge to a connected graph keeps it connected, and solutions to this problem can be solved using the approach from [59]. In light of the results from [59], it can be easily verified that the methods from [33], [9] produce connected graphs, even tough they do not explicitly discuss it. Other admissible graph properties such as acyclic, $k$-partite or graphs with $k$-connected components with $k > 1$ are not closed under edge addition operations. Thus, leading to non-convex optimization problems. In addition to non-convexity, some graph families induce prohibitively large constraint sets, for instance there are $n^{n-2}$ different tree topologies with $n$ vertices.

---

**Algorithm 1** Graph learning with admissible graph families

**Require:** $\mathbf{S}$ and $\mathcal{F}$.
1: Initialization: construct the set $\tilde{\mathcal{E}}_0 = \{(i, j) : s_{ij} > 0\}$.
2: Graph topology inference (GTI): find an edge set $\tilde{\mathcal{E}} \subset \tilde{\mathcal{E}}_0$ such that $(\mathcal{V}, \tilde{\mathcal{E}}) \in \mathcal{F}$.
3: Graph weight estimation (GWE): find $\mathbf{L}^{\#}$ by solving

$$\min_{\mathbf{L} \in \mathbb{L}_n^+(\tilde{\mathcal{E}})} \mathcal{J}(\mathbf{L}).$$

---

Noting that since the solution of unconstrained minimization of $\mathcal{J}(\mathbf{L})$ is $\mathbf{S}^{-1}$, our approach can be interpreted as finding an approximate inverse of $\mathbf{S}$ that belongs to the set $\{\mathbf{L} \in \mathbb{L}_n^+ : \mathcal{G}(\mathbf{L}) \in \mathcal{F}\}$. For a fixed $\mathbf{S}$, we can classify Problem 1 into three types of graph learning problems.

1) $\mathbf{S}^{-1} \in \mathbb{L}_n^+$ and $\mathcal{G}(\mathbf{S}^{-1}) \in \mathcal{F}$: In this case we can solve Problem 1 by inverting $\mathbf{S}$. This scenario is highly unlikely, as $\mathbf{S}$ would in general be computed from data. An interesting variation of this scenario is analyzed in the papers for learning tree structured graphs from Table I. When the data follows a GMRF and the input is an empirical covariance matrix $\mathbf{S}$, a maximum weight spanning tree type of algorithm [40] is consistent, i.e. as the number of samples goes to infinity, it will infer the correct tree structure with probability 1.
2) $\mathbf{S}^{-1} \in \mathbb{L}_n^+$ and $\mathcal{G}(\mathbf{S}^{-1}) \notin \mathcal{F}$: This case leads to a graph modification problem, where given a graph and its GGL matrix, we want to find its optimal approximation in the desired graph family. The approximation problem can be interpreted as minimizing the Kullback-Leibler divergence between attractive GMRFs, or minimizing the log-determinant divergence between GGL matrices [38].
3) $\mathbf{S}^{-1} \notin \mathbb{L}_n^+$ and $\mathcal{G}(\mathbf{S}^{-1}) \notin \mathcal{F}$: This case corresponds to graph learning with a desired topology property. This scenario reflects better a real world situation, where $\mathbf{S}$ is computed from data, and the modeling assumptions do not hold.

We are more interested in finding approximate solutions for Problem 1 in the second and third scenarios. We also focus on deterministic approximation leaving statistical analysis for future work.

### D. Proposed algorithm

Since Problem 1 is non-convex, we propose Algorithm 1 for approximating its solution. The initialization step is required in order to achieve our theoretical guarantees, since any feasible solution will not include edges that have a non-positive similarity value (see Theorem 1). The graph topology inference (GTI) step is a generic routine that returns a feasible graph topology, i.e., an edge set that belongs to $\mathcal{F}$. We will go into more details on how to solve this step in Section VI. The graph weight estimation (GWE) step is a convex optimization problem.

In this paper, we deal with three optimization problems, which are Problems 1, 2 and the GWE step from Algorithm 1. The proof of our error bound for Algorithm 1 (Theorem 4), relies on three results that can be stated informally as

1) The GWE step from Algorithm 1 can be replaced by Problem 2 with a specific choice of regularization $\mathbf{\Gamma}$ (Proposition 2), so that regularization with $\mathbf{\Gamma}$ can be used instead of sparsity constraints.

2) We derive necessary conditions on the regularization matrix $\mathbf{\Gamma}$ and the similarity matrix $\mathbf{S}$ guaranteeing that the solution of Problem 2 will have an admissible graph topology (Proposition 3), which will allow us to choose $\mathbf{\Gamma}$ for our problem.

3) Under certain conditions, we can use the solution of Problem 2, a convex problem, to approximate the solution of the non-convex Problem 1 (Lemma 1 and Theorem 4).

## IV. ANALYSIS OF PROBLEM 2

The main results of this section are Theorems 1 and 2, which characterize the relation between the regularization parameters, the similarity matrix and the graph of the solution of Problem 2. Later, in Proposition 2 we show the equivalence between the GWE step and Problem 2. In Proposition 3 we show how the solution Problem 2 is related to admissible graph families. We end this section with a formula for the solution of Problem 2 for acyclic graphs.

We begin by simplifying Problem 2. Throughout this section, the regularization matrix $\mathbf{\Gamma} = (\gamma_{ij})$ can be arbitrary symmetric, non-negative, with diagonal entries equal to zero ($\gamma_{ii} = 0$). Since $\mathbf{L}$ is a GGL matrix, it has non positive off diagonal entries, hence the $\ell_1$ regularization term from Problem 2 can be written as $- \operatorname{tr}(\mathbf{\Gamma L})$. Then, Problem 2 has a more compact form given by

$$\min_{\mathbf{L} \in \mathbb{L}_n^+} - \log\det(\mathbf{L}) + \operatorname{tr}(\mathbf{KL}),$$

with $\mathbf{K} = \mathbf{S} - \mathbf{\Gamma}$. We now introduce the covariance thresholding graph.

**Definition 6.** Given $\mathbf{K} = \mathbf{S} - \mathbf{\Gamma}$, the covariance thresholding graph is defined as $\tilde{\mathcal{G}}(\mathbf{K}) = (\mathcal{V}, \tilde{\mathcal{E}}, \mathbf{A})$ with connectivity matrix $\mathbf{A} = (a_{ij})$ corresponding to

$$a_{ij} = 1 \text{ if } k_{ij} > 0, \text{ and } i \neq j$$
$$a_{ij} = 0 \text{ otherwise.}$$

Since $k_{ij} = s_{ij} - \gamma_{ij}$, edges are included in the covariance thresholding graph when the similarity is larger than the regularization parameter.

Problem 2 is convex so that the *Karush Kuhn Tucker* (KKT) conditions are necessary and sufficient for optimality [35]. We denote by $\mathbf{L}^{\#} = (l_{ij}^{\#})$ the solution of Problem 2, with corresponding graph $\mathcal{G}(\mathbf{L}^{\#}) = (\mathcal{V}, \mathcal{E}^{\#}, \mathbf{L}^{\#})$. The optimality conditions are:

$$-(\mathbf{L}^{\#})^{-1} + \mathbf{K} + \mathbf{\Lambda} = 0 \qquad (2)$$
$$\mathbf{\Lambda} = \mathbf{\Lambda}^T \geq \mathbf{0} \qquad (3)$$
$$\operatorname{diag}(\mathbf{\Lambda}) = \mathbf{0} \qquad (4)$$
$$\mathbf{\Lambda} \odot \mathbf{L}^{\#} = \mathbf{0} \qquad (5)$$
$$\forall i \neq j, l_{ij}^{\#} \leq 0 \qquad (6)$$
$$\mathbf{L}^{\#} \succ 0, \qquad (7)$$

where $\mathbf{\Lambda} = (\lambda_{ij})$ is a matrix of Lagrange multipliers and $\odot$ is the Hadamard (entry-wise) product.

**Remark 1.** *The edge set of the covariance thresholding graph $\tilde{\mathcal{E}}$, and the graph of the optimal solution of Problem 2 $\mathcal{G}(\mathbf{L}^{\#})$, use the same notation from Algorithm 1. In later sections we will show this equivalence for a particular choice of regularization matrix $\mathbf{\Gamma}$.*

### A. Sparsity of optimal graphs

In this section we show that some of the edges of the optimal GGL matrix can be removed by screening the entries of the input matrix $\mathbf{K}$.

**Theorem 1** (Sparsity). *Given $\tilde{\mathcal{G}}(\mathbf{K}) = (\mathcal{V}, \tilde{\mathcal{E}}, \mathbf{A})$ and $\mathcal{G}(\mathbf{L}^{\#}) = (\mathcal{V}, \mathcal{E}^{\#}, \mathbf{L}^{\#})$ as defined before, we have that $\mathcal{E}^{\#} \subset \tilde{\mathcal{E}}$.*

*Proof.* We have to show that if $l_{ij}^{\#} < 0$, then $k_{ij} > 0$. By complementary slackness, $l_{ij}^{\#} < 0$ implies that $\lambda_{ij} = 0$. Then, based on Lemma 2 (see Appendix A), we can write $(\mathbf{L}^{\#})_{ij}^{-1} = k_{ij}$ as

$$((\mathbf{L}^{\#})^{-1})_{ij} = \frac{-l_{ij}^{\#}}{l_{ii}^{\#} l_{jj}^{\#}} + e_{ij} = k_{ij}.$$

Thus, we have $k_{ij} > 0$ since $e_{ij} \geq 0$. $\qquad \square$

Theorem 1 provides a lot of information about the solution of Problem 2 by only looking at the covariance thresholding graph of $\mathbf{K}$. Some implications of Theorem 1 are the following:

- The number of edges in the optimal graph is bounded by the number of edges in the covariance thresholding graph, i.e., $|\mathcal{E}^{\#}| \leq |\tilde{\mathcal{E}}|$.
- The optimal graph structure is partially known. If $k_{ij} \leq 0$, or equivalently if the similarity is below a threshold, the optimal graph does not include the corresponding edge. The entries of the regularization matrix act as thresholds for the similarity matrix.
- The converse is not necessarily true, if $k_{ij} > 0$ we cannot say if the optimal weight $l_{ij}^{\#}$ is zero or not. We will show that for acyclic graphs we can exactly determine the graph topology (see Theorem 3).

Although the above theorem is a simple result with a straightforward proof, it is the key technical element that connects the results of this paper and justifies the use of Algorithm 1. To the best of our knowledge, there is no counterpart to Theorem 1 for graphical Lasso or other inverse covariance estimation algorithm.

### B. Connected components of optimal graph

In this section we show that the optimal graph and the covariance thresholding graph have the same connected components. We can decompose $\tilde{\mathcal{G}}(\mathbf{K}) = (\mathcal{V}, \tilde{\mathcal{E}}, \mathbf{A})$ into its connected components $\{\tilde{\mathcal{G}}_s(\mathbf{K})\}_{s=1}^{J}$. Each $\tilde{\mathcal{G}}_s(\mathbf{K}) = (\tilde{\mathcal{V}}_s, \tilde{\mathcal{E}}_s, \mathbf{A}_s)$ is a connected sub-graph and $\{\tilde{\mathcal{V}}_1, \cdots, \tilde{\mathcal{V}}_J\}$ forms a partition of the nodes $\mathcal{V}$. Also for different components $\tilde{\mathcal{V}}_s, \tilde{\mathcal{V}}_t$ with $s \neq t$, if we pick $i \in \tilde{\mathcal{V}}_s$, and $j \in \tilde{\mathcal{V}}_t$ there is no edge connecting them (i.e., $a_{ij} = 0$).

Similarly, consider the solution of Problem 2 and its corresponding graph $\mathcal{G}(\mathbf{L}^\#) = (\mathcal{V}, \mathcal{E}^\#, \mathbf{L}^\#)$. Let $\{\mathcal{G}_s(\mathbf{L}^\#)\}_{s=1}^M$ denote its connected components defined by $\mathcal{G}_s(\mathbf{L}^\#) = (\mathcal{V}_s, \mathcal{E}_s^\#, \mathbf{L}_{\mathcal{V}_s}^\#)$, where $\mathbf{L}_{\mathcal{V}_s}^\#$ is a $|\mathcal{V}_s| \times |\mathcal{V}_s|$ sub-matrix of $\mathbf{L}^\#$ that only keeps rows and columns indexed by the set $\mathcal{V}_s$. We state the following theorem on connected components obtained by solving Problem 2.

**Theorem 2.** *If $\mathbf{L}^\#$ is the solution of Problem 2, then the connected components of $\mathcal{G}(\mathbf{L}^\#)$ and $\tilde{\mathcal{G}}(\mathbf{K})$ induce the same vertex partition, i.e. $M = J$ and there is a permutation $\pi$, such that $\tilde{\mathcal{V}}_s = \mathcal{V}_{\pi(s)}$ for $s \in [J]$.*

*Proof.* See Appendix B. □

This result has the following consequence.

**Corollary 1.** *$\tilde{\mathcal{G}}(\mathbf{K})$ is connected if and only if $\mathcal{G}(\mathbf{L}^\#)$ is connected.*

This allows us to check whether the solution is connected by only checking if the covariance thresholding graph is connected, without solving Problem 2.

### C. Regularization and sparsity

Since Theorem 1 informs us about some of the zero entries of $\mathbf{L}^\#$, then solving Problem 2 is equivalent to solving

$$\min_{\mathbf{L} \in \mathbb{L}_n^+(\tilde{\mathcal{E}})} -\log\det(\mathbf{L}) + \mathrm{tr}(\mathbf{K}\mathbf{L}), \tag{8}$$

where $\tilde{\mathcal{E}}$ is the edge set of the covariance thresholding graph. In particular we have that some instances of Problem 2 can be replaced by a non regularized sparsity constrained problem.

**Proposition 2.** *Let $\tilde{\mathcal{E}}$ be an arbitrary edge set, for example the one produced by the GTI step in Algorithm 1. If the regularization matrix is constructed as*

$$\gamma_{ij} = \begin{cases} 0 & \text{if } (i,j) \in \tilde{\mathcal{E}} \\ s_{ij} & \text{otherwise.} \end{cases} \tag{9}$$

*Then $\mathbf{L}^\#$ minimizes Problem 2 and also*

$$\mathbf{L}^\# = \arg\min_{\mathbf{L} \in \mathbb{L}_n^+(\tilde{\mathcal{E}})} -\log\det(\mathbf{L}) + \mathrm{tr}(\mathbf{S}\mathbf{L}).$$

*In particular, the GGL estimation step of Algorithm 1 can be replaced by Problem 2.*

*Proof.* This choice of regularization implies that $k_{ij} = s_{ij}$ for $(i,j) \in \tilde{\mathcal{E}}$, and $k_{ij} = 0$ otherwise. From Theorem 1 we have that edges are not allowed outside $\tilde{\mathcal{E}}$, thus we can add the constraint $\mathbf{L} \in \mathbb{L}_n^+(\tilde{\mathcal{E}})$ without changing the solution. □

This result indicates that thresholding some entries of the covariance matrix can be replaced by constraining the corresponding entries of the GGL matrix to be zeros. We can also use Theorem 2 to simplify Problem 2. The rows and columns of matrices $\mathbf{A}$ and $\mathbf{L}$ can be permuted to have the same block diagonal structure, thus they can be written as

$$\mathbf{L} = \begin{pmatrix} \mathbf{L}_{\mathcal{V}_1} & 0 & \cdots & 0 \\ 0 & \mathbf{L}_{\mathcal{V}_2} & \cdots & 0 \\ \vdots & \vdots & \ddots & \vdots \\ 0 & 0 & \cdots & \mathbf{L}_{\mathcal{V}_J} \end{pmatrix}$$

and

$$\mathbf{A} = \begin{pmatrix} \mathbf{A}_{\mathcal{V}_1} & 0 & \cdots & 0 \\ 0 & \mathbf{A}_{\mathcal{V}_2} & \cdots & 0 \\ \vdots & \vdots & \ddots & \vdots \\ 0 & 0 & \cdots & \mathbf{A}_{\mathcal{V}_J} \end{pmatrix}.$$

This block diagonal structure can be inferred by inspecting $\mathbf{A}$, and the optimal GGL matrix will have the same block diagonal structure. We can restrict ourselves to consider GGL matrices with the block structure of $\mathbf{A}$. The determinant of a block diagonal matrix is the product of the determinants of the individual blocks. Since the trace is linear on $\mathbf{L}$, we can decompose (8) into $J$ smaller sub-problems, one for each connected component, and solve

$$\min_{\mathbf{L} \in \mathbb{L}_{n_s}^+(\tilde{\mathcal{E}}_s)} -\log\det(\mathbf{L}) + \mathrm{tr}(\mathbf{K}_{\mathcal{V}_s}\mathbf{L}) \tag{10}$$

$\forall s \in \{1, \cdots, J\}$, where $n_s = |\mathcal{V}_s|$. Thus, solving Problem 2 is equivalent to solving $J$ smaller problems of size $n_s$ by only screening the values of $\mathbf{K}$ and identifying the connected components of $\tilde{\mathcal{G}}(\mathbf{K})$. For larger regularization parameter, the covariance thresholding graph will become sparse and possibly disconnected, thus affecting the sparsity and connected components of the solution of Problem 2. Since each of the $J$ GGL estimation problems are independent, they can be solved in parallel and asynchronously using any of the algorithms from [13], [33], [34], [49].

### D. Closed form solution for acyclic graphs

Acyclic graphs form a special graph family that admits closed form solutions. First note that combining Theorems 1 and 2, implies that the number of edges in the optimal graph satisfies

$$-J + \sum_{s=1}^J |\mathcal{V}_s| \leq |\mathcal{E}^\#| \leq |\tilde{\mathcal{E}}|, \tag{11}$$

where $J$ is the number of connected components. These inequalities are sharp when the graph $\tilde{\mathcal{G}}(\mathbf{K})$ is acyclic. We state this more precisely in the following Theorem.

**Theorem 3.** *If $\tilde{\mathcal{G}}(\mathbf{K})$ is an acyclic graph, then the solution of Problem 2 has the same edge set, i.e., $\mathcal{E}^\# = \tilde{\mathcal{E}}$, and the optimal GGL matrix is given by*

$$l_{ij}^\# = \begin{cases} \frac{1}{k_{ii}}\left(1 + \sum_{j \in \mathcal{N}_i} \frac{k_{ij}^2}{k_{ii}k_{jj} - k_{ij}^2}\right) & i = j \\ -\frac{k_{ij}}{k_{ii}k_{jj} - k_{ij}^2} & (i,j) \in \mathcal{E}^\# \\ 0 & (i,j) \notin \mathcal{E}^\#. \end{cases} \tag{12}$$

*Proof.* From Theorem 1 we have that $\mathcal{E}^\# \subset \tilde{\mathcal{E}}$. Since the covariance thresholding graph is acyclic, and from Theorem 2 both graphs have the same number of connected components, then (11) holds with equality. Given that both edge sets have the same number of elements, and one of them is included in the other, they must be equal. The authors of [39] showed that GGL matrices (which they call M-matrices), have a closed form expression in terms of their inverse. Denote $\boldsymbol{\Sigma} = \mathbf{L}^{-1}$, and note that the KKT optimality conditions establish that $\Sigma_{ij} = k_{ij}$ if $i = j$ or $(i, j) \in \mathcal{E}^\#$. The formula in (12) follows directly from [39, Corollary 2.3]. □

The same formula was obtained in [54] for the case when the graphical Lasso solution is a tree. The main difference with our approach is that we use results for GGL matrices of tree structured graphs that lead to a simple proof. The work of [39] is in the context of graph theory and does not make the connection with inverse covariance estimation methods.

### E. Admissible graph families

The theory developed thus far allows us to prove the following proposition.

**Proposition 3** (Transitive property)**.** *Let $\mathcal{F}$ be an admissible graph family, if $\tilde{\mathcal{G}}(\mathbf{K}) \in \mathcal{F}$, then the solution of Problem 2, $\mathcal{G}(\mathbf{L}^*) = (\mathcal{V}, \mathcal{E}^\#, \mathbf{L}^*)$, belongs to $\mathcal{F}$.*

*Proof.* The case of monotone graph properties follows directly from Theorem 1. The case of graphs with $k$-connected components follows from Theorem 2. When the graph class is an intersection of a monotone family and the family of graphs with $k$-connected components, the property follows again by applying Theorems 1 and 2. □

## V. ANALYSIS OF ALGORITHM 1

This section is devoted to proving Theorem 4, which deals with the output of Algorithm 1. We also obtain a more refined result for acyclic graphs. A GGL that minimizes Problem 1 is denoted by $\mathbf{L}^*$, and its graph by $\mathcal{G}(\mathbf{L}^*) = (\mathcal{V}, \mathcal{E}^*, \mathbf{L}^*)$. Since Problem 1 is non-convex, $\mathbf{L}^*$ might not be the only global minimum.

Our first result is Lemma 1 below, which states that Problem 1 can be approximated by Problem 2.

**Lemma 1.** *Let $\mathbf{L}^\#$ be the solution of Problem 2, with corresponding graph $\mathcal{G}(\mathbf{L}^\#) = (\mathcal{V}, \mathcal{E}^\#, \mathbf{L}^\#)$. Suppose that*

$$\mathcal{G}(\mathbf{L}^\#) \in \mathcal{F}. \tag{13}$$

*Then the following inequality holds*

$$|\mathcal{J}(\mathbf{L}^\#) - \mathcal{J}(\mathbf{L}^*)| \leq \sum_{i \neq j} \gamma_{ij} w_{ij}^*,$$

*where $|l_{ij}^*| = -l_{ij}^* = w_{ij}^*$ for all $i \neq j$*

*Proof.* Since $\mathbf{L}^\#$ is optimal for Problem 2 we have that

$$\mathcal{J}(\mathbf{L}^\#) + \sum_{i \neq j} \gamma_{ij} |l_{ij}^\#| \leq \mathcal{J}(\mathbf{L}^*) + \sum_{i \neq j} \gamma_{ij} |l_{ij}^*|. \tag{14}$$

Optimality of $\mathbf{L}^*$ for Problem 1 and (13) implies that

$$\mathcal{J}(\mathbf{L}^*) \leq \mathcal{J}(\mathbf{L}^\#).$$

We conclude by combining both inequalities

$$\sum_{i \neq j} \gamma_{ij} |l_{ij}^*| \geq \mathcal{J}(\mathbf{L}^\#) - \mathcal{J}(\mathbf{L}^*) + \sum_{i \neq j} \gamma_{ij} |l_{ij}^\#|$$
$$\geq \mathcal{J}(\mathbf{L}^\#) - \mathcal{J}(\mathbf{L}^*) \geq 0.$$

□

We denote the weight of a sub-graph of the covariance/similarity graph as

$$\mathcal{W}_{\mathbf{S}}(\mathcal{E}) = \sum_{(i,j) \in \mathcal{E}} s_{ij},$$

where $\mathcal{E}$ is an arbitrary edge set. The following Theorem is the main contribution of this paper.

**Theorem 4.** *Let $\mathcal{F}$ be an admissible graph family, and assume $\mathbf{S} \geq 0$. Let $\mathbf{L}^\#$ be the output of Algorithm 1, then we have that*
- *$\mathcal{G}(\mathbf{L}^\#) = (\mathcal{V}, \mathcal{E}^\#, \mathbf{L}^\#) \in \mathcal{F}$, and*
- *$|\mathcal{J}(\mathbf{L}^\#) - \mathcal{J}(\mathbf{L}^*)| \leq 2\bar{w} \{ \mathcal{W}_{\mathbf{S}}(\mathcal{E}_{full}) - \mathcal{W}_{\mathbf{S}}(\tilde{\mathcal{E}}) \}$,*

*where $|l_{ij}^*| = -l_{ij}^* = w_{ij}^*$ for all $i \neq j$, and $\bar{w} = \max_{(i,j) \notin \tilde{\mathcal{E}}} w_{ij}^*$. Also, $\tilde{\mathcal{E}}$ is the edge set found in the GTI step.*

*Proof.* In the previous section we used the matrix $\mathbf{K} = \mathbf{S} - \boldsymbol{\Gamma}$ to define the covariance thresholding graph. First we take the edge set $\tilde{\mathcal{E}}$ from the GTI step, and we choose $\boldsymbol{\Gamma}$ such that the covariance thresholding graph is $\tilde{\mathcal{G}}(\mathbf{K}) = (\mathcal{V}, \tilde{\mathcal{E}})$. This means we need to pick $\boldsymbol{\Gamma}$ such that $\gamma_{ij} < s_{ij}$ for $(i, j) \in \tilde{\mathcal{E}}$, and $\gamma_{ij} \geq s_{ij}$ when $(i, j) \notin \tilde{\mathcal{E}}$. Since the GTI step ensures that $\tilde{\mathcal{E}}$ belongs to $\mathcal{F}$, this choice of regularization combined with Proposition 3 imply that $\mathcal{G}(\mathbf{L}^\#) \in \mathcal{F}$. Now we can apply Lemma 2 and obtain the bound

$$|\mathcal{J}(\mathbf{L}^\#) - \mathcal{J}(\mathbf{L}^*)| \leq \sum_{i \neq j} \gamma_{ij} w_{ij}^*.$$

The goal now is to simplify the upper bound and make it as tight as possible. We start by decomposing the right hand side using the terms corresponding to $\tilde{\mathcal{E}}$ and $\tilde{\mathcal{E}}^c$

$$\sum_{i \neq j} \gamma_{ij} w_{ij}^* = 2 \sum_{(i,j) \in \tilde{\mathcal{E}}} \gamma_{ij} w_{ij}^* + 2 \sum_{(i,j) \notin \tilde{\mathcal{E}}} \gamma_{ij} w_{ij}^*$$
$$= 2 \sum_{(i,j) \notin \tilde{\mathcal{E}}} s_{ij} w_{ij}^*$$
$$\leq 2\bar{w} \sum_{(i,j) \notin \tilde{\mathcal{E}}} s_{ij},$$

where for the second equality we have chosen

$$\gamma_{ij} = \begin{cases} 0 & \text{if } (i, j) \in \tilde{\mathcal{E}} \\ s_{ij} & \text{otherwise.} \end{cases}$$

This choice produces the tightest possible upper bound because $\tilde{\mathcal{E}}$ is a well defined covariance thresholding graph, i.e., $0 \leq \gamma_{ij} < s_{ij}$ when $(i, j) \in \tilde{\mathcal{E}}$, and $s_{ij} \leq \gamma_{ij}$ when $(i, j) \notin \tilde{\mathcal{E}}$. For the last inequality we use

$$\bar{w} = \max_{(i,j) \notin \tilde{\mathcal{E}}} w_{ij}^*.$$

We conclude by applying Proposition 2 to Problem 2 with the chosen regularization, thus making the solution of Problem 2 coincide with the GWE step from Algorithm 1. □

Theorem 4 indicates that Problem 2, which is convex, can be close to the non-convex Problem 1 if the upper bound can be minimized. This can be achieved in several ways, including:

- If $\mathcal{E}^* \subset \tilde{\mathcal{E}}$ and $\tilde{\mathcal{E}}$ is in $\mathcal{F}$, then $\bar{w} = 0$. This is in general not possible, but in Section VI we will discuss a case where our method is equivalent to a consistent estimator of the graph structure, thus achieving $\bar{w} \to 0$.

- Most of the time finding a feasible graph that contains the optimal graph is not possible, hence $\bar{w} > 0$ but since $\tilde{\mathcal{E}}$ depends solely on the similarity matrix $\mathbf{S}$, maximizing $\mathcal{W}_{\mathbf{S}}(\mathcal{E})$ becomes a practical alternative. We will discuss this and other options to solve the graph topology inference step in Section VI.

The following proposition states a tighter error bound for acyclic graphs.

**Proposition 4.** *If $\mathcal{F}$ is the family of acyclic graphs. Under the same assumptions of Theorem 4, the solution of Algorithm 1 satisfies*

$$|\mathcal{J}(\mathbf{L}^\#) - \mathcal{J}(\mathbf{L}^*)| \leq 2 \sum_{(i,j) \in \mathcal{E}^* \cap \tilde{\mathcal{E}}^c} \frac{r_{ij}^2}{1 - r_{ij}^2} \leq 2 \sum_{(i,j) \notin \tilde{\mathcal{E}}} \frac{r_{ij}^2}{1 - r_{ij}^2},$$

*where $r_{ij} = s_{ij}/\sqrt{s_{ii}s_{jj}}$.*

*Proof.* We start from Lemma 1 and follow the same steps of Theorem 4, except that instead of bounding the entries $w_{ij}^*$, we use the closed form expression from Theorem 3. We replace the values for $w_{ij}^*$ and get

$$\begin{aligned}
|\mathcal{J}(\mathbf{L}^\#) - \mathcal{J}(\mathbf{L}^*)| &\leq 2 \sum_{(i,j) \in \mathcal{E}^* \cap \tilde{\mathcal{E}}^c} w_{ij}^* s_{ij} \\
&= 2 \sum_{(i,j) \in \mathcal{E}^* \cap \tilde{\mathcal{E}}^c} \frac{s_{ij}^2}{s_{ii}s_{jj} - s_{ij}^2} \\
&= 2 \sum_{(i,j) \in \mathcal{E}^* \cap \tilde{\mathcal{E}}^c} \frac{r_{ij}^2}{1 - r_{ij}^2} \\
&\leq 2 \sum_{(i,j) \notin \tilde{\mathcal{E}}} \frac{r_{ij}^2}{1 - r_{ij}^2}.
\end{aligned}$$

$\square$

## VI. Graph topology inference

In this section we propose two methods to solve the graph topology inference step of Algorithm 1. The first one is derived by analyzing the upper bound in Theorem 4. The second one is similar, but uses a normalized similarity matrix. We also show that for specific instances of graph families, the proposed graph topology inference methods can be solved using existing combinatorial optimization algorithms.

Minimizing the upper bound from Theorem 4 leads to a maximum weight spanning sub-graph approximation problem in $\mathcal{F}$ given by

$$\max_{(\mathcal{V},\mathcal{E}) \in \mathcal{F}} \mathcal{W}_{\mathbf{S}}(\mathcal{E}). \qquad (A_1)$$

For some graph families, solving $(A_1)$ might not be computationally tractable which affects the quality of the output of

Algorithm 1. We can see this by decomposing the second factor of the upper bound in Theorem 4 as:

$$\underbrace{\mathcal{W}_{\mathbf{S}}(\mathcal{E}_{full}) - \mathcal{MW}_{\mathbf{S}}^{\mathcal{F}}}_{modeling\ error} + \underbrace{\mathcal{MW}_{\mathbf{S}}^{\mathcal{F}} - \mathcal{W}_{\mathbf{S}}(\tilde{\mathcal{E}})}_{approximation\ error},$$

where and $\mathcal{MW}_{\mathbf{S}}^{\mathcal{F}} = \max_{(\mathcal{V},\mathcal{E}) \in \mathcal{F}} \mathcal{W}_{\mathbf{S}}(\mathcal{E})$, and $\tilde{\mathcal{E}}$ is the output of the GTI step, i.e. an exact or approximate solution of $(A_1)$. The *modeling error* measures how much the covariance matrix differs from its best approximation in $\mathcal{F}$. This term depends on the choice of graph family only and cannot be improved once $\mathcal{F}$ has been decided. The *approximation error* measures the discrepancy between the best approximation of the covariance by a thresholded matrix with a non-zero pattern in $\mathcal{F}$ and the estimate given by $\tilde{\mathcal{E}}$. This term depends on the existence of efficient algorithms for solving $(A_1)$. For example, when $\mathcal{F}$ is the family of bipartite graphs, solving $(A_1)$ is equivalent to max-cut, which is NP-hard, hence the approximation error will always be positive. When $\mathcal{F}$ is the family of tree structured graphs, the solution can be found exactly in polynomial time using Kruskal's algorithm, hence the only term is the modeling error. When $(A_1)$ is an NP-hard problem, good solutions with small approximation errors can be obtained using combinatorial approximation algorithms.

If we look at the graph learning problem as finding an approximate inverse matrix for $\mathbf{S}$, i.e. find $\mathbf{L}^\# \simeq \mathbf{S}^{-1}$, where the GGL matrix has a desired type of non-zero pattern, normalization on the entries of $\mathbf{S}$ can be beneficial as will be shown later in the experimental section.

As an intuitive reason for the use of normalization matrices, assume $\mathbf{S}$ has very unbalanced magnitude values, and the graph structure is very regular. During step 2 of Algorithm 1, more importance will be given to larger similarity values. However, for correct graph topology identification, all edges are equally important.

Based on this observation, we also use a normalized similarity matrix for the GTI step and solve

$$\max_{(\mathcal{V},\mathcal{E}) \in \mathcal{F}} \mathcal{W}_{\mathbf{R}}(\mathcal{E}) = \max_{(\mathcal{V},\mathcal{E}) \in \mathcal{F}} \sum_{(i,j) \in \mathcal{E}} \frac{s_{ij}}{\sqrt{s_{ii}s_{jj}}}, \qquad (A_2)$$

where $\mathbf{R} = \mathrm{diag}(\mathbf{S})^{-1/2} \mathbf{S} \, \mathrm{diag}(\mathbf{S})^{-1/2}$.

In the following subsections we will show specific instances of the GTI step for the families of tree structured graphs, $k$-sparse connected graphs and bipartite graphs. We will describe the method only for $(A_1)$, with the understanding that the same derivation applies for $(A_2)$.

### A. Tree structured graphs

The set of tree structured graphs is the set of $(n-1)$-sparse connected graphs. Theorem 4 suggests solving $(A_1)$, which reduces to the following problem for trees,

$$\max_{\mathcal{E}} \sum_{(i,j) \in \mathcal{E}} s_{ij} \text{ s.t. } |\mathcal{E}| \leq n - 1, (\mathcal{V}, \mathcal{E}) \text{ is connected.} \quad (15)$$

Moreover, Proposition 4 suggests solving the following optimization

$$\max_{\mathcal{E}} \sum_{(i,j) \in \mathcal{E}} \frac{r_{ij}^2}{1 - r_{ij}^2} \text{ s.t. } |\mathcal{E}| \leq n-1, (\mathcal{V}, \mathcal{E}) \text{ is connected.} \tag{16}$$

Both correspond to maximum weight spanning tree (MWST) problems (with different edge weights) which can be solved using Kruskal's or Prim's algorithms [60] in $\mathcal{O}(n^2 \log(n))$ time. For this graph family, the optimization problem exactly solves the combinatorial problem (approximation error is zero). When all the $r_{ij}$ are different, $(A_2)$ and (16) have the same solution, furthermore, they are equivalent to the Chow-Liu algorithm [40], formally stated in the following proposition.

**Proposition 5.** *Suppose the vectors $\{\mathbf{x}_i\}_{i=1}^N$ (columns of data matrix) are i.i.d. realizations of an attractive GMRF distribution, and all the values $r_{ij}$ are different for $i < j$. Then the GTI step of Algorithm 1 solved using $(A_2)$, the Chow-Liu algorithm and (16) have the same solution.*

*Proof.* The Chow-Liu algorithm [40] solves a MWST problem using the pairwise empirical mutual information as edge weights. For two Gaussian variables the mutual information is

$$I(\mathbf{x}^i, \mathbf{x}^j) = -\frac{1}{2} \log(1 - r_{ij}^2) = I_{ij},$$

where $r_{ij} = s_{ij}/\sqrt{s_{ii}s_{jj}}$. The mutual information is a non-negative strictly increasing function of $|r_{ij}|$, then any MWST algorithm will return the same tree structured graph for weights $|r_{ij}|$ and $I_{ij}$. Since all the $r_{ij}$ are non-negative, the graph topology inference step that uses $A_2$ and the Chow-Liu algorithm are equivalent. The same argument can be used to show that (16) and $(A_2)$ are equivalent, since $r^2/(1 - r^2)$ is an increasing function of $r^2$. $\square$

If the data has a tree structured distribution, the Chow-Liu algorithm is consistent [41], i.e. $\tilde{\mathcal{E}} \to \mathcal{E}^*$ with probability one as $N \to \infty$. From Theorem 3 we know that $\mathcal{E}^{\#} = \tilde{\mathcal{E}}$, then $\mathcal{E}^{\#} \to \mathcal{E}^*$ with probability one [41]. Therefore for this case, Algorithm 1 is consistent.

### B. Bipartite graphs

For the bipartite graph case, we need to find a partition of the vertex set $\mathcal{V} = \mathcal{S} \cup \mathcal{S}^c$, and an edge set $\tilde{\mathcal{E}}$ for which $(i,j) \in \tilde{\mathcal{E}}$ if and only if $i \in \mathcal{S}$ and $j \in \mathcal{S}^c$. Finding a bipartite approximation is equivalent to a max-cut problem between $\mathcal{S}$ and $\mathcal{S}^c$. A standard way of solving max-cut, is introducing a variable $y_i \in \{+1, -1\}$ for each node in the graph, where $y_i = +1$ if $i \in \mathcal{S}$ and $y_i = -1$ otherwise. This results in the optimization problem

$$\max_{y_1, \cdots, y_n} \sum_{i,j} s_{ij}(1 - y_i y_j) \text{ s.t. } y_i \in \{+1, -1\}. \tag{17}$$

Max-cut is NP hard, so approximation algorithms are the only option. We use the Goemans-Williamson (GM) algorithm [61], which finds a 0.87856-approximation for (17). The complexity of the GM algorithm is dominated by a semi-definite convex optimization problem and a Cholesky decomposition of an $n \times n$ matrix (see [61] for complexity analysis). Obviously, other approximation algorithms can be used for bipartite approximation with lower computational complexity. Once the vector $\mathbf{y} = [y_1, \cdots, y_n]$ is found, we construct the edge set as

$$\tilde{\mathcal{E}} = \{(i,j) : y_i y_j = -1\}.$$

### C. k-sparse connected graphs

For a fixed and known $k$ we want to solve

$$\max_{\mathcal{E}} \sum_{(i,j) \in \mathcal{E}} s_{ij} \text{ s.t. } |\mathcal{E}| \leq k, (\mathcal{V}, \mathcal{E}) \text{ is connected.} \tag{18}$$

When $k = n - 1$, the solution of (18) is the MWST. Based on this observation, for the case $k \geq n$ we propose the following approximation algorithm

1) Find the edge set $\mathcal{E}_0$ that solves the MWST from (15).
2) Find $\mathcal{E}_1$ with edges $(i,j)$ corresponding to the entries $s_{ij}$ with the $k-n+1$ largest magnitudes such that $(i,j) \notin \mathcal{E}_0$.
3) Return $\tilde{\mathcal{E}} = \mathcal{E}_0 \cup \mathcal{E}_1$.

The resulting graph is the maximum weight connected $k$-sparse graph that contains the MWST.

## VII. Experiments

In this section, we evaluate Algorithm 1 on synthetic and real data. To quantify the graph learning performance in terms of topology inference, we use the F-score (which is typically used for evaluating binary classification problems) defined as

$$\text{FS} = \frac{2\text{tp}}{2\text{tp} + \text{fp} + \text{fn}},$$

where tp, fp, fn correspond to *true positives, false positives* and *false negatives* respectively. The F-score takes values in $[0, 1]$ where a value of 1 indicates perfect classification. To evaluate GGL estimation, we use the relative error in Frobenius norm given by

$$\text{RE} = \frac{\|\mathbf{L}_{ref} - \mathbf{L}^{\#}\|_F}{\|\mathbf{L}_{ref}\|_F},$$

where $\mathbf{L}_{ref}$ is a reference GGL matrix, and $\mathbf{L}^{\#}$ denotes the output of Algorithm 1.

### A. Graph learning from synthetic data

In this section, we evaluate the performance of Algorithm 1 for learning graphs with three different topology properties. The main difficulty in evaluation of our algorithms is the fact that the optimal solution of Problem 1 cannot be found in general. Even designing an experiment with synthetic data and computing metrics to evaluate graph learning performance is non-trivial. Therefore, we will instead consider scenarios where we have an input empirical covariance matrix that is close to an attractive GMRF with a GGL matrix in the graph family $\mathcal{F}$. For this purpose, we generate data as i.i.d. realizations of the random variable

$$\mathbf{x} = \theta \mathbf{z}_1 + (1 - \theta)\mathbf{z}_2, \tag{19}$$

where $\theta \in \{0, 1\}$ is a Bernoulli random variable with probability $\mathbb{P}(\theta = 1) = \alpha$. The independent random variables $\mathbf{z}_1$ and $\mathbf{z}_2$ are distributed as zero-mean multivariate Gaussians $\mathcal{N}(\mathbf{0}, \mathbf{L}_{type}^{-1})$ and $\mathcal{N}(\mathbf{0}, \mathbf{L}_{ER}^{-1})$, respectively. The precision matrix of $\mathbf{z}_1$ is the GGL of a graph with an admissible graph topology $\mathcal{G}(\mathbf{L}_{type}) = (\mathcal{V}, \mathcal{E}_{type}, \mathbf{L}_{type})$, while the precision matrix of $\mathbf{z}_2$ is the GGL of an Erdos-Renyi random graph (which is not admissible). A simple calculation reveals that $\mathbf{x}$ is zero-mean with covariance matrix equal to

$$\mathbb{E}(\mathbf{x}\mathbf{x}^{\top}) = \boldsymbol{\Sigma} = \alpha\mathbf{L}_{type}^{-1} + (1 - \alpha)\mathbf{L}_{ER}^{-1}.$$

The inverse covariance matrix $\boldsymbol{\Sigma}^{-1}$ is not a GGL matrix, but for $\alpha \simeq 1$ we expect it to be well approximated by an attractive GMRF with the same graph topology as $\mathbf{L}_{type}$. In our experiments, we choose the inverse covariance matrix of $\mathbf{z}_2$ as the GGL of an Erdos-Renyi random graph, with probability of an edge of $p = 0.05$ and with weights sampled from the uniform distribution $U[1, 2]$. Note that, for a small choice of $\alpha$, $\mathbf{z}_2$ can be interpreted as a random noise model based on Erdos-Renyi graphs.

We compare our algorithms against two methods. First we consider a reference Laplacian matrix, obtained as

$$\mathbf{L}_{ref} = \underset{\mathbf{L} \in \mathbb{L}_n^+(\mathcal{E}_{type})}{\arg\min} -\log\det(\mathbf{L}) + \text{tr}(\mathbf{S}_N\mathbf{L}), \quad (20)$$

where $\mathbf{S}_N$ is the empirical covariance matrix obtained from $N$ i.i.d. realizations of the random vector $\mathbf{x}$. The constraint forces the topology of the reference GGL to be included in $\mathcal{E}_{type}$, and possibly having different graph weights, except when $N \to \infty$ and $\alpha = 1$. This method assumes that for smaller noise, i.e. $\alpha \simeq 1$, the correct graph topology should resemble $\mathcal{E}_{type}$.

Second, we compare against a *baseline* method that first learns a graph without any topology constraint using existing methods [33]. Then it simplifies the graph topology using the algorithms from Section VI on the learned graph. With the new topology, the GGL is estimated. We can summarize this procedure in the following steps:

1) Solve an unconstrained graph learning problem,

$$\mathbf{L}_{uc} = \arg\min_{\mathbf{L} \in \mathbb{L}_n^+} -\log\det(\mathbf{L}) + \text{tr}(\mathbf{S}_N\mathbf{L}), \quad (21)$$

   and $\mathcal{G}(\mathbf{L}_{uc}) \notin \mathcal{F}$.

2) Find graph topology

$$\mathcal{E}_{base} = \underset{(\mathcal{V}, \mathcal{E}) \in \mathcal{F}}{\arg\max} \sum_{(i,j) \in \mathcal{E}} |l_{ij}^{uc}|, \quad (22)$$

   where $\mathbf{L}_{uc} = (l_{ij}^{uc})$.

3) Find graph weights

$$\mathbf{L}_{base} = \underset{\mathbf{L} \in \mathbb{L}_n^+(\mathcal{E}_{base})}{\arg\min} -\log\det(\mathbf{L}) + \text{tr}(\mathbf{S}_N\mathbf{L}). \quad (23)$$

Our theoretical analysis indicates that step 1) in the aforementioned procedure can be removed, and step 2) can be solved using the covariance matrix. In this section, we experimentally validate our theoretical analysis by demonstrating that our proposed methods outperform the baseline method.

*1) Learning tree structured graphs:* We sample 50 uniform random trees[2] with $n = 50$ vertices. Each edge in the graph is assigned a weight drawn from an uniform distribution $U[1, 2]$, while the vertex weights are drawn from a $U[0, 1]$. For each graph and each $N$ (number of i.i.d. realizations), we run 50 Monte-Carlo simulations. The proposed methods are compared against the reference Laplacian obtained by solving (20) and the baseline method. In Fig. 1a, we plot the F-score with respect to the reference Laplacian as a function of the parameter $N/n$ (i.e., number of observed data samples per vertex). As expected, our algorithms find graph topologies similar to that of the reference Laplacian, and for smaller $\alpha$, the topologies become more dissimilar. Using $(A_2)$ to solve the graph topology inference step results in better performance for all values of $N/n$ and $\alpha$, which also provides the best performance in graph topology estimation with smaller number of samples. Even though the graph topologies are close (i.e., F-score values are similar), the graph weights can be significantly different, as reflected in the relative error and objective function values from Figures 1b and 1c. Similarly, we observe that solving $(A_2)$ leads to solutions that are closer to the reference Laplacian. For larger number of samples, the relative error and objective functions converge, where the output of Algorithm 1 and the reference Laplacian become very close. Even though $\mathbf{x}$ is a Gaussian mixture, $(A_2)$ is still better than $(A_1)$, while it is no longer equivalent to the Chow-Liu algorithm.

The baseline method is slightly better than $(A_2)$ for $N/n \leq 5$, while for larger values of $N$ the performance becomes indistinguishable. This is remarkable, since the baseline method estimates the graph topology based on the solution of an unconstrained GGL estimation problem, while our proposed GTI method only uses the sample covariance matrix.

*2) Learning bipartite graphs:* We construct GGL matrices $\mathbf{L}_{type}$ of bipartite graphs with $n = 50$ vertices as follows. First we partition the vertices into $\mathcal{S} = \{1, \cdots, 25\}$ and its complement $\mathcal{S}^c$. Then, the edges that connect $\mathcal{S}$ with $\mathcal{S}^c$ are added with probability $q = 0.5$. Finally, the edge weights and vertex weights are drawn from uniform distribution $U[1, 2]$ and $U[0, 1]$ respectively. We run Algorithm 1 and use the max-cut approximation from [61] to solve the GTI step. To evaluate graph learning performance, we report F-scores for the vertex partition, relative errors and value of the objective function, all with respect to the optimal GGL matrix found by solving (20). We observe that normalization, i.e., solving $(A_2)$, is helpful but not as much as for trees, and the performances of solving $(A_1)$ and $(A_2)$ are almost the same in terms of all three metrics. The F-scores are smaller relative to tree case, but they improve with more data samples $N/n$, suggesting the obtained graph topologies are not substantially different to the reference graph. However, notice that the reference Laplacian, i.e., the solution of (20), has higher cost than the output of our algorithms, for all values of $N/n$ and $\alpha$, as seen in Figure 2c.

For $\alpha = 0.95$, the baseline method is outperformed by all other methods when $N/n \leq 10$. A similar behavior is observed for $\alpha = 0.75$, but only when $N/n \leq 100$. As expected, the values of the objective function verify that the baseline method is slightly better than all other methods, suggesting that the reference graph topology is not optimal when $N/n$ is smaller.

---

[2]All trees of $n$ nodes are picked with the same probability given by $1/n^{n-2}$. We generate them by constructing Prüfer sequences.

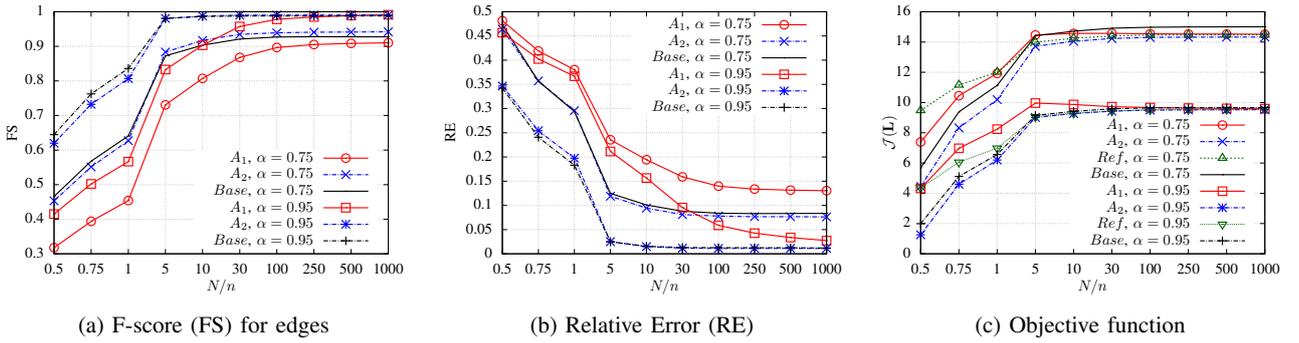

Fig. 1: Performance of Algorithm 1 with graph topology found using $(A_1)$ and $(A_2)$ for learning tree structured graphs.

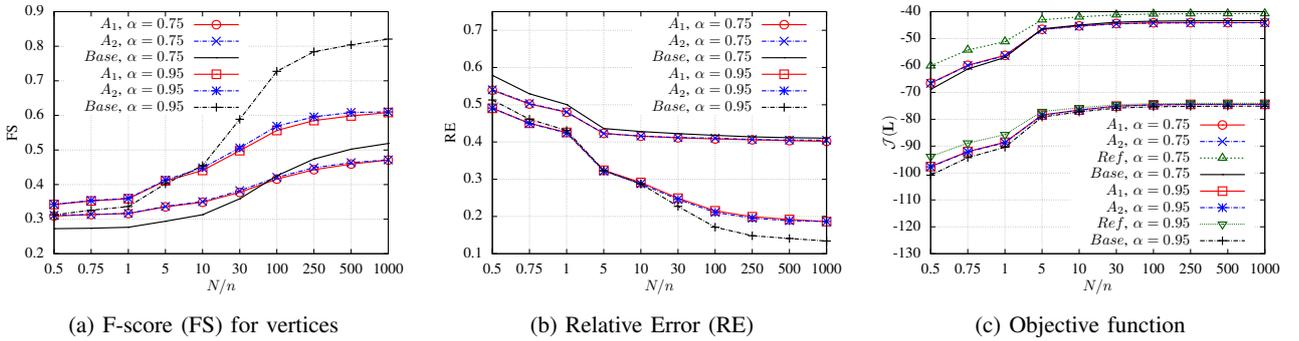

Fig. 2: Performance of Algorithm 1 with graph topology found using $(A_1)$ and $(A_2)$ for learning bipartite graphs.

*3) Connected $k$-sparse graphs:* We construct connected $k$-sparse graphs as follows. First we generate an uniform random tree with $n = 50$ vertices, then we randomly connect pairs of vertices until there are $k$ edges. The GGL matrix $\mathbf{L}_{type}$ is then constructed by assigning random edge and vertex weights drawn from the uniform distributions $U[1, 2]$ and $U[0, 1]$, respectively. We choose $k = 150$ (sparsity of $\approx 12\%$) for all experiments. We follow the same procedure as before and compare the reference GGL against the output of Algorithm 1 with GTI found by solving $(A_1)$ and $(A_2)$. With respect to $\mathbf{L}_{ref}$, we observe from Figures 3a and 3b that normalization produces better topologies for all parameters, although the gain is larger for small $N/n$. However, we observe from Figure 3c that the graph topology given by the reference graph is not optimal, since Algorithm 1 always produce GGL matrices with smaller cost. When compared to the baseline method, we observe that for $\alpha = 0.75$ the performance is almost as good as $(A_2)$. For $\alpha = 0.95$, the baseline method and $(A_2)$ show identical performance until $N/n > 1$, when the proposed methods $(A_1)$ and $(A_2)$ start improving in F-score and RE. An interesting observation is that, for very large $N/n$, the baseline method and the output of Algorithm 1 have different graph topologies, as indicated by the F-Score. However, they all converge to the same objective function value. This is due to the non-convexity of the problem, since multiple solutions with the same cost value may exist.

### B. Image graphs

In this section, we evaluate Algorithm 1 with image data. We use the Brodatz texture images from the USC-SIPI dataset[3]. We take the subset of rotated textures *straw\*.tiff* and partition each image into non-overlapping blocks of size $8 \times 8$. Each block is vectorized and organized as columns of the data matrix $\mathbf{X}$. Then, we use Algorithm 1 to learn (i) tree, (ii) bipartite and (iii) connected, sparse graphs. For trees and connected sparse graphs, we use the sample covariance matrix $\mathbf{S}_N$. For bipartite graphs, we use a thresholded sample covariance matrix $\mathbf{S}_N \odot \mathbf{A}$, where $\mathbf{A}$ is the 0-1 connectivity matrix of a 8-connected grid graph [1]. In Figure 4, we show the resulting graphs for three *straw* textures. The edge weights are colored and have been normalized to lie in $[0, 1]$. The resulting graphs all have strong link weights that follow the texture orientation. In particular for trees, the strong weights connect pixels along the main texture orientation, while the weaker edge weights connect pixels in other directions. For bipartite graphs, the vertices (i.e., pixels) are colored in red and black, to mark the bipartition obtained during the graph topology inference step of Algorithm 1.

Neighboring pixels along the direction of the texture orientation have strong edges, and at the same time neighboring pixels have different color, providing a reasonable and intuitive notion of *every other pixel*. For sparse connected graphs, even though we choose the same parameter $k = 112$ for all images, the number of connections in the optimal graph is always smaller. The graphs corresponding to *straw000*, *straw060* and *straw150* images have 105, 109 and 107 edges, respectively.

---

[3]http://sipi.usc.edu/database/database.php?volume=textures

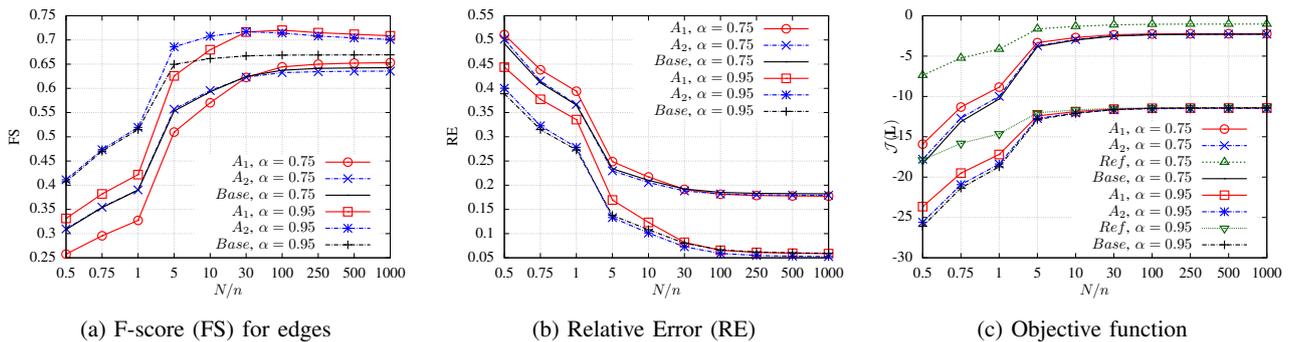

(a) F-score (FS) for edges     (b) Relative Error (RE)     (c) Objective function

Fig. 3: Performance of Algorithm 1 with graph topology found using ($A_1$) and ($A_2$) for learning connected $k$-sparse graphs.

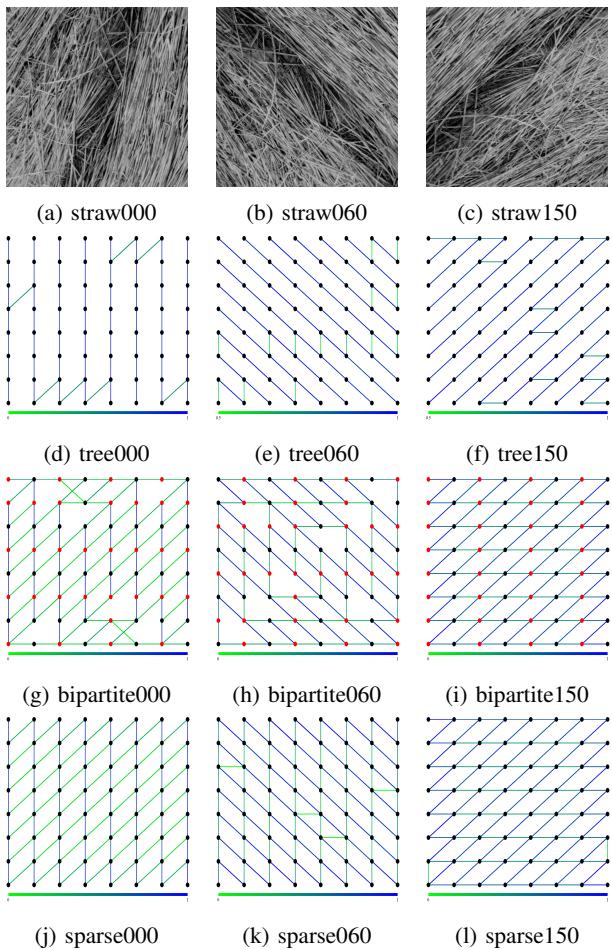

(a) straw000    (b) straw060    (c) straw150

(d) tree000    (e) tree060    (f) tree150

(g) bipartite000    (h) bipartite060    (i) bipartite150

(j) sparse000    (k) sparse060    (l) sparse150

Fig. 4: Structured graphs learned from sample covariance matrices of texture images using Algorithm 1 with normalized graph topology inference ($A_2$). (a)-(c) $512 \times 512$ texture images. (d)-(f) tree structured graphs. (g)-(i) bipartite graphs. (j)-(l) sparse connected graphs obtained by setting sparsity parameter $k = 112$.

## VIII. CONCLUSION

We have proposed a framework to learn graphs with admissible topology properties. These include monotone graph families, graphs with a given number of connected components, and graph families that are in the intersection of both. The proposed algorithm consists of two main steps. First, a graph topology with the desired property is found by solving a maximum weight spanning sub-graph problem on the similarity matrix. And second, a generalized graph Laplacian matrix is found by solving a graph topology constrained log-determinant divergence minimization problem. Although finding the best graph with the desired property is in general a non-convex optimization problem, we show that our proposed solution is near optimal in the sense that seeks to minimize an error bound with respect to the best possible solution. Our theoretical guarantees are based on the analysis of a convex relaxation of the non-convex problem and relies on properties of generalized graph Laplacian matrices. We implemented instances of our algorithm to learn tree structured graphs, bipartite graphs and sparse connected graphs, and tested them with synthetic and texture image data. Our theoretical and numerical results indicate that one can find a near optimal graph topology by screening the entries of the covariance/similarity matrix.

# APPENDIX

## A. Properties of GGL matrices

The following property of GGL matrices will be used throughout the paper.

**Lemma 2.** *Let* $\mathbf{L} = \mathbf{V} + \mathbf{D} - \mathbf{W}$ *be a positive definite GGL, and denote* $\mathbf{P} = \mathrm{diag}(\mathbf{L}) = \mathbf{V} + \mathbf{D}$. *Then*

$$\mathbf{L}^{-1} = \mathbf{P}^{-1} + \mathbf{P}^{-1}\mathbf{W}\mathbf{P}^{-1} + \mathbf{E} \qquad (24)$$

*for some* $\mathbf{E} \geq 0$.

*Proof.* First decompose $\mathbf{L}$ as

$$\mathbf{L} = \mathbf{P} - \mathbf{W} = \mathbf{P}^{1/2}(\mathbf{I} - \mathbf{Q})\mathbf{P}^{1/2},$$

where $\mathbf{Q} = \mathbf{P}^{-1/2}\mathbf{W}\mathbf{P}^{-1/2}$. Since $\mathbf{L}$ is positive definite, then $\mathbf{I} - \mathbf{Q}$ is also positive definite hence $\|\mathbf{Q}\|_2 < 1$, so we have the following set of equalities:

$$\begin{aligned}
\mathbf{L}^{-1} &= \mathbf{P}^{-1/2}(\mathbf{I} - \mathbf{Q})^{-1}\mathbf{P}^{-1/2} \\
&= \mathbf{P}^{-1/2}(\mathbf{I} + \mathbf{Q} + \mathbf{Q}^2 + \cdots)\mathbf{P}^{-1/2} \\
&= \mathbf{P}^{-1} + \mathbf{P}^{-1}\mathbf{W}\mathbf{P}^{-1} + \mathbf{E}
\end{aligned}$$

where $\mathbf{E} = \mathbf{P}^{-1/2}(\sum_{k=2}^{\infty}\mathbf{Q})\mathbf{P}^{-1/2} \geq 0$. □

A consequence is the following.

**Theorem 5.** *[58] Let* $\mathbf{L} = \mathbf{V} + \mathbf{D} - \mathbf{W}$ *be a GGL matrix, then the following statements are equivalent*

1) $\mathbf{L}$ *is non singular and* $\mathbf{L}^{-1} \geq 0$.
2) $\mathbf{L} \in \mathbb{L}_n^+$.

When $\mathbf{L}$ is an inverse covariance matrix, Theorem 5 implies that attractive GMRF models have non-negative covariance matrices.

## B. Proof of Theorem 2

*Proof.* We need to show that after a permutation of rows and columns, the matrices $\mathbf{A}$ and $\mathbf{L}^{\#}$ as defined before, have the same block diagonal structure. Since we need to prove a set equality, we will show inclusion in one direction and then the converse.

($\Rightarrow$) First assume that the rows and columns of $\mathbf{A}$ have been reordered so the matrix is block diagonal. Each block is indexed by the vertex partition given by the connected components of $\tilde{\mathcal{G}}(\mathbf{K})$. Pick two connected components $s \neq t$ and $i \in \mathcal{V}_s, j \in \mathcal{V}_t$, hence $(i,j) \notin \tilde{\mathcal{E}}$ and $a_{ij} = 0$ therefore $k_{ij} \leq 0$. Using Theorem 1 we have $l_{ij}^{\#} = 0$, and since this is true for all $i \in \mathcal{V}_s, j \in \mathcal{V}_t$ then $(i,j) \notin \mathcal{E}^{\#}$, there are not any edges between vertices in $\mathcal{V}_s$ and $\mathcal{V}_t$ for the graph $\mathcal{G}(\mathbf{L}^{\#})$. Then there exist $s' \neq t'$ such that $\tilde{\mathcal{V}}_{s'} \subset \mathcal{V}_s$ and $\tilde{\mathcal{V}}_{t'} \subset \mathcal{V}_t$. Since this is true for all $s, t, i, j$ we also have that there are at least $J$ connected components in $\mathcal{G}(\mathbf{L}^{\#})$, i.e. $M \geq J$.

($\Leftarrow$) For the converse the proof is similar, again assume the rows and columns of $\mathbf{L}^{\#}$ have been reordered so the matrix is block diagonal with blocks given by the vertices of the connected components of $\mathcal{G}(\mathbf{L}^{\#})$. Then pick two connected components $s \neq t$ and $i \in \hat{\mathcal{V}}_s, j \in \hat{\mathcal{V}}_t$, hence $(i,j) \notin \mathcal{E}^{\#}$ and $l_{ij}^{\#} = 0$. Since $\mathbf{L}^{\#}$ is block diagonal, then $(\mathbf{L}^{\#})^{-1}$ is also block diagonal therefore $((\mathbf{L}^{\#})^{-1})_{ij} = 0 = k_{ij} + \lambda_{ij}$, where the second equality comes from (3). Using (3) then $\lambda_{ij} \geq 0$, and $k_{ij} \leq 0$, which implies $(i,j) \notin \tilde{\mathcal{E}}$ and $a_{ij} = 0$. Using the same argument as before, we have that there exist $s'' \neq t''$ such that $\mathcal{V}_{s''} \subset \hat{\mathcal{V}}_s$ and $\mathcal{V}_{t''} \subset \hat{\mathcal{V}}_t$ and $J \geq M$.

We have shown that $J = M$, now using ($\Rightarrow$) fix any $s$, then there exists an unique $s'$ (because $L = M$) such that $\tilde{\mathcal{V}}_{s'} \subset \mathcal{V}_s$, now using ($\Leftarrow$) there also exists an unique $s''$ such that $\mathcal{V}_{s''} \subset \hat{\mathcal{V}}_{s'}$. Since the $\{\mathcal{V}_s\}$ are disjoint the only possibility is that $s'' = s$ which implies $\mathcal{V}_s = \hat{\mathcal{V}}_{s'}$ ans since the mapping is one on one, we have that the permutation is given by $\pi(s) = s'$. □